\documentclass[conference]{IEEEtran}
\IEEEoverridecommandlockouts
\usepackage{cite}
\usepackage{amsmath,amssymb,amsfonts}
\usepackage{algorithmic}
\usepackage{graphicx}
\usepackage{textcomp}
\usepackage{xcolor}
\usepackage{setspace}

\usepackage{algorithmic}
\usepackage{booktabs}  
\usepackage{threeparttable} 
\usepackage{multirow}
\usepackage{booktabs}
\usepackage[misc]{ifsym}
\usepackage{hyperref}
\hypersetup{hidelinks,
	colorlinks=true,
	allcolors=black,
	pdfstartview=Fit,
	breaklinks=true}
\def\BibTeX{{\rm B\kern-.05em{\sc i\kern-.025em b}\kern-.08em
		T\kern-.1667em\lower.7ex\hbox{E}\kern-.125emX}}

\begin{document}

\title{TieFake: Title-Text Similarity and Emotion-Aware Fake News Detection 
	\thanks{\dag \quad Corresponding Author: Zhao Kang, zkang@uestc.edu.cn}
 }
\author{\IEEEauthorblockN{Quanjiang Guo$^1$, Zhao Kang$^1$\textsuperscript{\dag}, Ling Tian$^1$, Zhouguo Chen$^2$}
	\IEEEauthorblockA{$^1$School of Computer Science and Engineering, University of Electronic Science and Technology of China, Chengdu, China}
	\IEEEauthorblockA{$^2$30$^{th}$ Research Institute of China Electronics Technology Group Corporation, Chengdu, China}
        \IEEEauthorblockA{guochance1999@163.com, zkang@uestc.edu.cn, lingtian@uestc.edu.cn, czgexcel@163.com }}

\maketitle

\begin{abstract}
Fake news detection aims to detect fake news widely spreading on social media platforms, which can negatively influence the public and the government. Many approaches have been developed to exploit relevant information from news images, text, or videos. However, these methods may suffer from the following limitations: (1) ignore the inherent emotional information of the news, which could be beneficial since it contains the subjective intentions of the authors; (2) pay little attention to the relation (similarity) between the title and textual information in news articles, which often use irrelevant title to attract reader’ attention. To this end, we propose a novel \textbf{\underline{Ti}}tle-Text similarity and \textbf{\underline{e}}motion-aware \textbf{\underline{Fake}} news detection (\textbf{TieFake}) method by jointly modeling the multi-modal context information and the author sentiment in a unified framework. Specifically, we respectively employ BERT and ResNeSt to learn the representations for text and images, and utilize publisher emotion extractor to capture the author's subjective emotion in the news content. We also propose a scale-dot product attention mechanism to capture the similarity between title features and textual features. Experiments are conducted on two publicly available multi-modal datasets, and the results demonstrate that our proposed method can significantly improve the performance of fake news detection. Our code is available at \href{https://github.com/UESTC-GQJ/TieFake}{https://github.com/UESTC-GQJ/TieFake}.

\end{abstract}


\section{Introduction}
\label{sec:intro}
\begin{figure}[htbp]  
	\centering
	\includegraphics[width=80mm,scale=1.00]{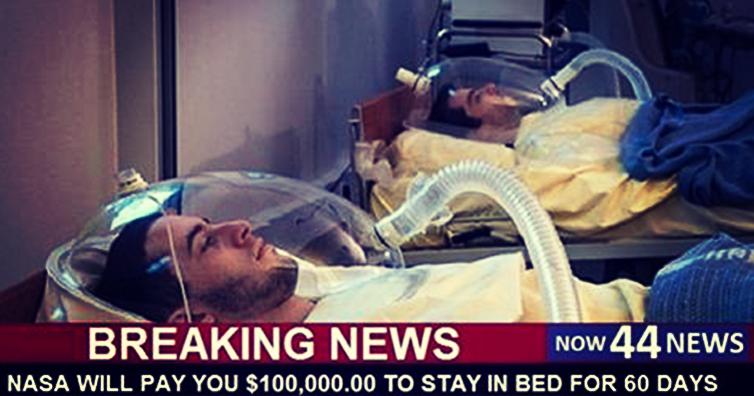}	
	\caption{An example of fake news that claims that NASA will pay subject \$100,000 to stay in bed for 60 days (politifact.com).}
	\label{fig:example1}
\end{figure}

\begin{figure}[htbp]  
	\centering
	\includegraphics[width=80mm,scale=1.00]{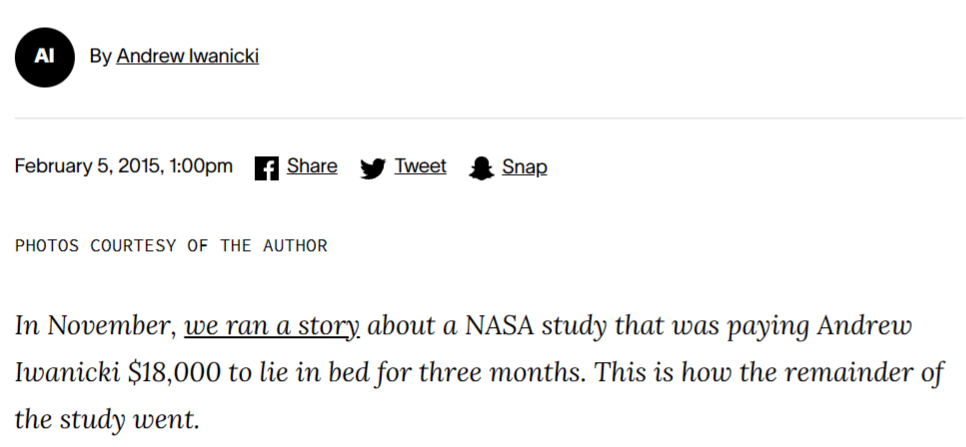}	
	\caption{The fact stated by the experimenter.}
	\label{fig:example2}
\end{figure}

Nowadays, more and more people consume news through social media. A recent study \cite{allcott2017} defines fake news “to be news articles that are intentionally and verifiably false, and could mislead readers.” Moreover, such content is written to deceive someone. An example of such fake news is shown in Figure \ref{fig:example1}. The image shown in the news is obviously photo-shopped to make it look similar to the news generally featured on a popular news channel like CNN. This image makes people believe that the news is true, but the experimenter himself quashed it (see Figure \ref{fig:example2}).

Especially since the 2016 U.S. presidential election, the dissemination of ``fake news" has adversely affected the public and the government. Based on a thorough analysis of 126,000 verified real and fraudulent news on Twitter from 2006 to 2017,
% Based on a broad investigation of 126,000 verified true and fake news stories on Twitter from 2006 to 2017,
Vosoughi et al.~\cite{vosoughi2018spread} pointed out that fake news and inaccurate information may spread faster and broader than fact-based news. According to crucial psychological and social science ideas~\cite{zhou2020survey}, the more fake news articles circulate, the more likely social media users will disseminate and believe them due to repeated exposure or peer pressure.
% As indicated by the fundamental theories in psychology and social sciences~\cite{zhou2020survey}, the more a fake news article spreads, the higher the possibility of social media users spread and trust it due to repeated exposure or peer pressure.
Due to the \emph{echo chamber effect}, such levels of trust and beliefs can easily be magnified and sustained within social media~\cite{jamieson2008echo}.
% Such levels of trust and beliefs can easily be amplified and reinforced within social media because of echo chamber effect~\cite{jamieson2008echo}.
Therefore, to prevent the dissemination of fake news on social media, extensive research has been done on the effective identification of fake news. 
%Hence, extensive research has been conducted on the effective detection of fake news to block its dissemination on social media
\begin{figure*}[ht]
    \centering
    \includegraphics[width=172mm]{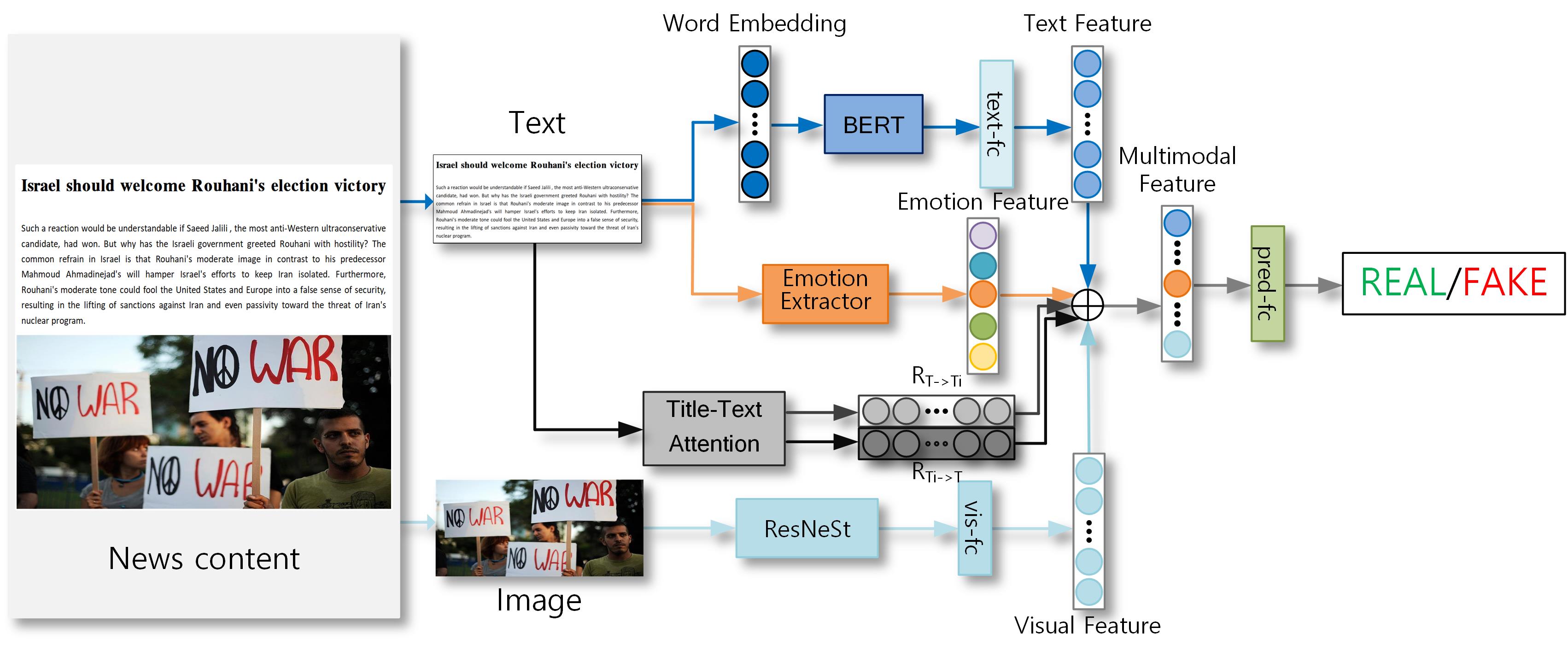}
    \caption{Overview of the $\mathsf{TieFake}$ framework. The method consists of five modules, (1) textual feature extraction (Sec.\ref{subsec:Textual Feature Extractor}); (2) visual feature extraction (Sec.\ref{subsec:Visual Feature Extractor}); (3) emotional feature extraction (Sec.\ref{subsec:Emotion Feature Extractor}); (4) Title-Text similarity  extraction (Sec.\ref{subsec:Title-text similarity extractor}); (5) multi-modal features fusion and fake news prediction (Sec.\ref{subsec:Fake news detector}).}
    \label{fig:framework}
\end{figure*}

Fake news detection methods can be roughly grouped into content-based and social-context-based strategies~\cite{shu2017fake}. The main difference between them is whether or not they rely on social context information: the information about how the news has been propagated on social media, where many auxiliary details of social media users involved and their connections/networks can be utilized. Many novel approaches~\cite{qian2018neural,DAI2022107659,fang2022structurepreserving,liuhongfei} have been proposed to exploit social context and social node information with GNN network. Nevertheless, it is often difficult to detect fake news using these methods when it has been just published and has not been propagated, i.e., no social context information, which motivates us to further explore the news content information in fake news detection.

A news article often contains both textual and visual information. Existing content-based fake news detection methods either solely consider textual information~\cite{zhou2020fake} or sentiment information~\cite{ajao2019sentiment,giachanou2019leveraging}, or combine information derived from both modalities, which complement each other in detecting fake news~\cite{wang2018eann,jin2016novel,jin2017multimodal}. Though their promising performance, they ignore the latent emotion of authors conveyed by fake news. To attract public attention, the authors prefer to show emotions with strong subjective color whose emotion is far from that in general statements of the news text~\cite{newman2003lying}. The authors also attempt to catch the reader's attention by using titles that are not relevant to the contents \cite{shrestha2021}.
In light of such considerations, we propose a \textbf{\underline{Ti}}tle-Text similarity and \textbf{\underline{e}}motion-aware \textbf{\underline{Fake}} news detection (\textbf{TieFake}) method. Figure \ref{fig:framework} shows the overview architecture of our proposed method. For each news article, we first adopt neural networks to automatically obtain the latent representation of both its textual and visual information. In addition, the subjective emotion of authors are extracted to help predict fake news. We also apply a scaled dot-product attention on title and text, because a huge “gap” always exists between the title and text of fake news when creators articulate or fabricate fake news to support non-factual scenes or statements. Therefore, the proposed method makes full use of the news text, images, the subjective emotion of author and the similarity between title and text to improve the accuracy of fake news detection.

The main contributions of our work are summarized as follows:
%\vspace{-1.0em}
\begin{itemize}
    \item To our best knowledge, we present the first approach that utilizes the subjective emotion of the news author in detecting fake news;
    %\vspace{-1.0em}
    \item We propose a novel attention mechanism to explore the similarity
between title and text;
    
  %  \item We propose a new framework to jointly exploit multi-modal (textual and visual) and emotional information of the author to learn the representation of news articles;
    %\vspace{-1.0em}
    \item We conduct extensive experiments on two real-world data to demonstrate the effectiveness of our proposed method.
    %\vspace{-1.0em}
\end{itemize}

\section{Related Work}
\label{sec:relatedwork}
\subsection{Fake News Detection}
Most existing work on fake news detection task treats it as a binary classification problem. Early methods~\cite{castillo2011information, ma2016detecting} design plenty of hand-crafted features to debunk fake news. These methods train a fake news classifier using text content features. Although these manually chosen features improve the performance of fake news detection,  these approaches typically require extensive preprocessing and feature engineering. Recognizing and detecting fake news has become more complex as social media information has exploded in popularity. Researchers have put forth various practical methods~\cite{castillo2011information, jin2017multimodal, kwon2013,shu2017fake, wang2018eann}, which can be briefly reviewed from single-modal (e.g., text or image) and multi-modal perspectives.

Existing methods\cite{castillo2011information, gupta2014, kwon2013, shu2017fake} for single-modal analysis mainly concentrate on extracting textual or visual elements from the news' text content or image. For example, Yu et al.~\cite{yu2017convolutional} use convolutional neural networks to obtain high-level interactions and critical features of related news. Recurrent neural networks are used by Ma et al.~\cite{ma2016detecting} to learn latent properties from the relevant textual news. In ~\cite{peng2019}, the authors only exploit the rich visual information with different pixel domains and utilize a multi-domain visual neural network to detect fake news. However, social media platforms offer a wealth of multi-modal data (e.g., images, texts, and videos)\cite{qian2016,qian2016article} that can complement each other and contribute to social media analysis\cite{song2020,wu2008}.

Researchers realize that multi-modal fusion features may be crucial in identifying fake news because of deep neural networks' enormous success in learning picture and word representations. Multi-modality fake news detection has recently drawn a lot of interest. Several approaches \cite{jin2017multimodal,khattar2019mvae,singhal2019spotfake,singhal2020spotfake+,wang2020,zhou2020safe} conduct fake news detection based on multimedia content and obtain superior performance. Jin et al.\cite{jin2017multimodal} propose a multi-modality-based fake news detection model, which extracts the multi-modality information, including visual, textual, and social context features, and then fuses them by attention mechanism. Khattar et al.\cite{khattar2019mvae} propose a multimodal variational autoencoder that learns a shared representation of the text and images. Shivangi et al.\cite{singhal2019spotfake} make use of the pre-trained BERT to learn text features and apply VGG-19 pre-trained on the ImageNet dataset to learn image features. Wang et al.\cite{wang2020} propose a novel knowledge-driven multimodal graph convolutional network to jointly model textual information, knowledge concepts, and visual information into a unified framework for fake news detection.

Although most existing approaches show promising performance on fake news detection task, they still fail to fully exploit the data. In this paper, we propose  to take advantage of the author's potential emotion and the similarity between the title and text.

\subsection{Attention Mechanism}
Attention mechanisms have been shown to be effective in various
tasks such as image captioning~\cite{xu2015,ren2022multi}, machine translation \cite{bahdanau2014} and recommendation system\cite{chen2017}. Bahdanau et al. \cite{bahdanau2014} firstly introduce attention to the machine translation task to allow the model to automatically search for parts of a source sentence that are relevant to predicting a target word. Soon after,  Transformer\cite{vaswani2017} is proposed to solve the sequence-to-sequence problem, replace LSTM with an attention structure, and achieve a state-of-the-art quality score on the neural machine translation task. Recently, attention mechanisms have been incorporated into fake news detection methods.
For example, Chen et al.\cite{chen2018} propose a deep attention model based on recurrent neural networks(RNN) to learn selectively temporal hidden representations of sequential posts for identifying fake news. Motivated by the successful applications of the attention mechanism, we introduce a scaled dot-product attention mechanism on title features and textual features to compute the similarity between news title and text.

\begin{figure*}[!htbp]
    \centering
    \includegraphics[width=165mm]{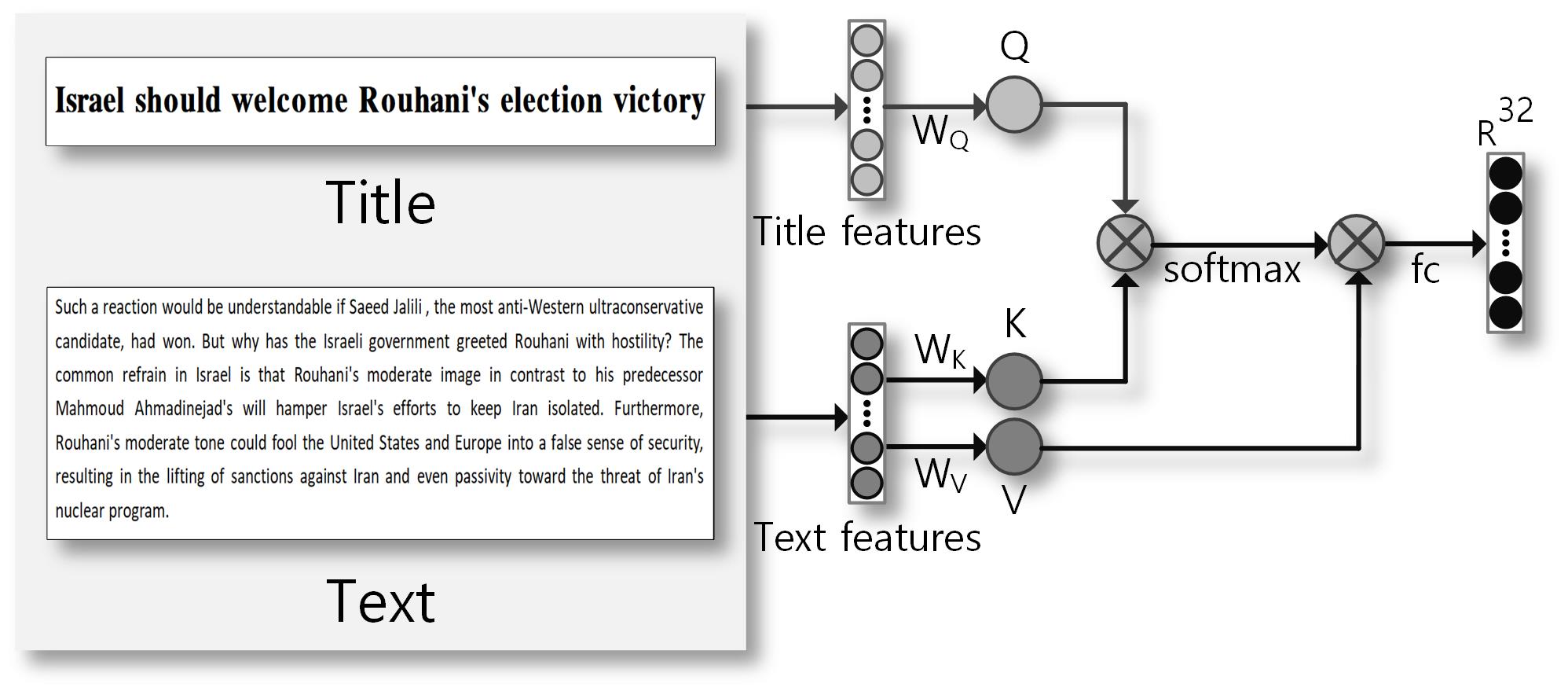}
    \caption{Title-Text attention}
    \label{fig:att}
\end{figure*}
\section{Proposed Method}
\label{sec:propose}
\noindent\textbf{\textit{Problem Definition}} Given a news article $A = \{t, v, e\}$ consisting of textual information $t$, visual information $v$, and emotion information $e$, we denote $\mathbf{R_T}\in\mathbb{R}^{d}$, $\mathbf{R_V} \in\mathbb{R}^{d}$, and $\mathbf{R_E} \in \mathbb{R}^{d}$ as the corresponding representations. Our goal is to predict whether $A$ is a fake news article ($\hat{y}=0$) or a true one ($\hat{y}=1$) by determining $\mathcal{M}: (\mathcal{M}_t,\mathcal{M}_v,\mathcal{M}_e) \xrightarrow{(\theta_t, \theta_v, \theta_e)} \hat{y} \in \{0,1\}$, where $\theta$ are parameters to be learned.

\subsection{Textual feature extractor}
\label{subsec:Textual Feature Extractor}
To precisely utilize the semantic meaning of the word and the linguistic contexts, we employ a 12 encoding layers version of Bidirectional Encoder Representations from Transformers (BERT~\cite{devlin2018bert}), which takes a sequence of words as input. We model a given text content $t$ as the textual embedding $R_{T_{bert}}$ = $\{\mathbf{T}_1, \mathbf{T}_2 , \cdots , \mathbf{T}_m\}$ (where $m$ denotes the number of words) from the second last output layer of the module, and passes the embedding through a fully-connected layer to reduce down to the final dimension of length 32, i.e., $R_T \in \mathbb{R}^{32}$. The operation of the fully-connected layer in the textual feature extractor can be represented as:
\begin{equation}
\label{eq:textual_matrix}
    R_T = \sigma (W_{vt} \cdot R_{T_{bert}})
\end{equation}
where $\sigma$ is the ReLU activation function and $W_{vt}$ is the weight matrix of the fully connected layer.
Similarly, we can use the same method to generate the feature vector of the title  $R_{Ti} \in \mathbb{R}^{32}$.

\subsection{Visual feature extractor}
\label{subsec:Visual Feature Extractor}
Given an image $V$ attached to a news text, we employ the pre-trained ResNeSt-50~\cite{zhang2022resnest}. On top of the last layer of ResNeSt-50, we add a fully connected layer to adjust the dimension of final visual feature representation to length 30, i.e., $R_V \in \mathbb{R}^{30}$. During the joint training process with the textual feature extractor, the parameters of pre-trained ResNeSt-50 are kept static to avoid overfitting. The operation of the last layer in the visual feature extractor can be represented as:
\begin{equation}
\label{eq:visual_matrix}
    R_V = \sigma (W_{vf} \cdot R_{V_{resnest}})
\end{equation}
where $R_{V_{resnest}}$ is the visual feature representation obtained from pre-trained ResNeSt-50 and $W_{vf}$ is the weight matrix of the fully connected layer in the visual feature extractor.

\subsection{Emotion feature extractor}
\label{subsec:Emotion Feature Extractor}
To comprehensively represent emotion, we employ publisher emotion extractor~\cite{zhang2021mining} to obtain a variety of features from news contents, including the emotion category, emotional lexicon, emotional intensity, sentiment score, and other auxiliary features. The corresponding feature representations are denoted by $R_{E}^{cate}$, $R_{E}^{lex}$, $R_{E}^{int}$, $R_{E}^{senti}$, and $R_{E}^{aux}$, respectively. Among them, emotion category, emotional intensity, and sentiment score provide the overall information, while the other two provide word- and symbol-level information.
We concatenate all five kinds of features as $R_E$ with final dimension of length 38, i.e.,
\begin{equation}
\label{eq:emotional_concate}
    R_E = R_{E}^{cate} \oplus R_{E}^{lex} \oplus R_{E}^{int} \oplus R_{E}^{senti} \oplus R_{E}^{aux}
\end{equation}

\subsection{Title-Text similarity extractor}
\label{subsec:Title-text similarity extractor}
In this part, we propose to apply the scaled dot-product attention mechanism on title features and textual features to capture how well the title and text are related to each other in the news. We utilize the attention mechanism for the title and textual features in two directions. Figure \ref{fig:att} shows the general design of our proposed attention mechanism.

When we use information from a text to compare to a title or use the text vector representation $R_T$ as Query and title features $R_{Ti}$ as Key and Value. Mathematically, the Query, Key, and Value are defined as:
\begin{equation}
\label{eq:similarity}
    \begin{aligned}
    Q &= R_{T} \times W_{Q}, 
    K = R_{Ti} \times W_{K},
    V = R_{Ti} \times W_{V}\\
    &W_{Q} \in \mathbb{R}^{d_{T} \times d},
    W_{K} \in \mathbb{R}^{d_{Ti} \times d},
    W_{V} \in \mathbb{R}^{d_{Ti} \times d},
    \end{aligned}
\end{equation}
where $d_{Ti}=d = 32$, $W_{Q}$, $W_{K}$, $W_{V}$ are weight matrices.

The output matrix of the scaled dot-product attention applied
on $Q$, $K$, and $V$ is calculated as:
\begin{equation}
\label{eq:att}
    Att_{T \xrightarrow{} Ti} = 
    softmax(\frac{Q \times K^{T}}{\sqrt{d}}) \times V.
\end{equation}
where $Att_{T \xrightarrow{} Ti}$ is the attention's output matrix when we use text vector representation $R_T$ as Query and title features $R_{Ti}$ as Key and Value.

Similarly, when we use information from the title to make comparison to the text, we obtain $Att_{Ti \xrightarrow{} T}$, in which the title vector representation $R_{Ti}$ is used as Query and $R_T$ is used as Key and Value. After obtaining two attention’s output matrices $Att_{T \xrightarrow{} Ti}$ and $Att_{Ti \xrightarrow{} T}$ , we pass them into a fully connected layer with the size of 32, which is the same
as $d_{T}$. Finally, we obtain two vectors $R_{T \xrightarrow{} Ti}$ and $R_{Ti \xrightarrow{} T}$.

\subsection{Fake news detector}
\label{subsec:Fake news detector}
We deploy a fully connected layer with softmax to predict whether the news articles are fake or real. Above five feature vectors $R_T$, $R_V$, $R_E$, $R_{T \xrightarrow{} Ti}$, and $R_{Ti \xrightarrow{} T}$ are fused together to generate the final news representation denoted as: 
\begin{equation}
\label{eq:detect}
    R_F =R_T \oplus R_V \oplus R_E \oplus R_{T \xrightarrow{} Ti} \oplus R_{Ti \xrightarrow{} T} \in \mathbb{R}^{164}
\end{equation}
where $\oplus$ is the concatenation operator. The fake news detector is built on top of the multi-modal feature extractors and takes $R_F$ as input. Finally, given a news article, its label $\hat{y}$ can be predicted by a fully connected layer with all learned parameters.

\section{EXPERIMENT AND Results}
\label{sec:exper}

\begin{table}[h]
\scriptsize
    \centering
	\setlength{\belowcaptionskip}{1pt}
	\setlength{\abovecaptionskip}{1pt}
    \newcommand{\tabincell}[2]{\begin{tabular}{@{}#1@{}}#2\end{tabular}}
    \centering
    \caption{Data statistics.}
    \vspace{+0.5em}
    \scalebox{1.4}{
    \begin{tabular}{cccc}
    \toprule
        \textbf{News articles} & \textbf{Fake} & \textbf{True} &  \textbf{Overall} \\
    \midrule
    \midrule
     \textbf{PolitiFact} & 161 & 205 & 366 \\
     \midrule
     
     \textbf{GossipCop}  & 4,927 & 16,693 & 21,620 \\
    \bottomrule
    \end{tabular}}
    \label{tab::datasets}
\end{table}

\subsection{Experimental setup}
\label{ssec:setup}
\noindent\textbf{\textit{Dataset}} 
We use two reputable public benchmark datasets from FakeNewsNet~\cite{shu2020fakenewsnet} for our experiments.
News articles are respectively collected from PolitiFact and GossipCop. PolitiFact (\url{politifact.com}) is a well-known non-profit fact-checking website of political statements and reports in the U.S. GossipCop (\url{gossipcop.com}) is a website that fact-checks celebrity reports and entertainment stories published in magazines and newspapers. The PolitiFact and GossipCop datasets contain news articles published between May 2002 to July 2018 and July 2000 to December 2018, respectively.
Experts in the relevant fields gave ground truth labels (\textit{fake} or \textit{true}), ensuring news labels' accuracy.
In this work, we focus on detecting fake news by incorporating text and image information. Thus, we remove news articles without any text or image, and statistics of the two datasets are provided in Table~\ref{tab::datasets}.

\noindent\textbf{\textit{Evaluation metrics}}
We commonly use Accuracy (Acc) as the
evaluation metric for binary classification tasks such as fake news
detection. However, its reliability is significantly compromised when a dataset suffers from class imbalance. Therefore, in addition to the Acc metric, Precision (Pre), Recall (Rec), and $F_1$ score are also deployed.

\noindent\textbf{\textit{Implementation details}} 
In our experiments, we split each dataset into training and testing sets in an 8:2 ratio. In our proposed model, we keep weights of pre-trained BERT and ResNeSt fixed and used them as feature extractors, because in preliminary experiments, we found that fine-tuning BERT and ResNeSt did not improve the performance of our model. Hyper-parameters used in experiments are as follows. The experiments are conducted on an AMD Ryzen 5800H CPU and an NVIDIA GeForce RTX 3060 GPU with 6GB RAM. Our algorithms are implemented on Pytorch deep learning framework\cite{Paszke2017AutomaticDI} and are trained with Adaptive Moment Estimation (Adam)\cite{Kingma2014AdamAM} optimizer. We use the same batch size of 16 instances in the training stage for all methods, and the model is trained for 10 epochs with a learning rate of $10^{-4}$.

\subsection{Baseline}
\label{ssec:baseline}

We compare to the following baselines, which detect fake news using (i) textual (Sec.\ref{subsubsec:textual}), (ii) visual (Sec.\ref{subsubsec:visual}), or (iii) multi-modal information (Sec.\ref{subsubsec:muti-modal}).
\subsubsection{Text-based models}
\label{subsubsec:textual}

\begin{itemize}
    \item \textbf{CNN}~\cite{yu2017convolutional}: CNN employs a convolutional neural network to learn the feature representations for misinformation identification and early detection tasks by framing news text into the fixed-length sequence;
    \item \textbf{GRU}~\cite{ma2016detecting}: GRU is based on recurrent neural networks (RNN) for learning the hidden representations that can use the multilayer GRU network to consider the post as a variable-length time series;
    \item \textbf{BERT}~\cite{devlin2018bert}: BERT is a bidirectional encoder from Transformer designed to pre-train deep bidirectional representations from unlabeled text with conditional computations common to both left and right contexts.
    % \item 
\end{itemize}

\subsubsection{Image-based models}
\label{subsubsec:visual}

\begin{itemize}
    \item \textbf{VGG-19}~\cite{simonyan2014very}: VGG-19 is a widely-used CNN with 19 layers for image classification. We use a fine-tuned VGG-19 as one of the baselines;
    \item \textbf{ResNet-50}~\cite{he2016deep}: ResNet-50 is a widely-used CNN with 50 layers in various feature extraction applications.
\end{itemize}

\subsubsection{Multi-modal models}
\label{subsubsec:muti-modal}
\begin{itemize}
    \item \textbf{MVAE}~\cite{khattar2019mvae}: MVAE uses a bimodal variational autoencoder coupled with a binary classifier for fake news detection;
    \item \textbf{att-RNN}~\cite{jin2017multimodal}: att-RNN is a deep neural network model applicable for multi-modal fake news detection. It employs LSTM and VGG-19 with attention mechanism to fuse news articles' textual, visual, and social-context features. We set the hyper-parameters the same and exclude the social-context features for a fair comparison;
    \item \textbf{SpotFake}~\cite{singhal2019spotfake}: SpotFake utilizes the pre-trained language models (like BERT) to learn the textual information and employs VGG-19 (pre-trained on the ImageNet dataset) to obtain image features;
    \item \textbf{EANN}~\cite{wang2018eann}: EANN can derive event-invariant features and thus assist in detecting fake news on newly arrived events, which consists of the multi-modal feature extractor, the fake news detector, and the post discriminator. For fairness of comparison, we conduct experiments with a simplified version of EANN that excludes the post discriminator;
    \item \textbf{SpotFake+}~\cite{singhal2020spotfake+}: SpotFake+ is an advanced version of SpotFake that extracts the textual feature using a pre-trained XLNet model;
    \item \textbf{SAFE}~\cite{zhou2020safe}: SAFE extracts multi-modal (textual and visual) features of news content and their relationships through a similarity-aware multi-modal method for fake news detection.
\end{itemize}

\begin{table*}[!htbp]
\scriptsize
    \centering
	\setlength{\belowcaptionskip}{1pt}
	\setlength{\abovecaptionskip}{1pt}
\centering
% \small
\caption{Performance of various methods in detecting fake news.}

\label{tab::general_performance}
% \begin{adjustbox}{\textwidth}
% \resizebox{\linewidth}{!}
\scalebox{1.5}{
\begin{threeparttable}
\setlength{\tabcolsep}{5mm}{
\begin{tabular}{|c|c|c|c|c|c|}
\hline
\textbf{Dataset}&
\textbf{Methods} &
\textbf{Acc} & 
\textbf{Pre} &
\textbf{Rec} & 
\textbf{$\mathbf{F}_1$}\\
\hline
\hline
 & \textbf{CNN} & 0.658 & 0.702 & 0.622 & 0.660 \\ 
 & \textbf{GRU} & 0.681 & 0.667 & 0.632 & 0.644 \\ 
 & \textbf{BERT} & 0.781 & 0.766 & 0.878 & 0.818 \\ \cline{2-6} 
 & \textbf{VGG-19} & 0.458 & 0.492 & 0.473 & 0.482 \\ 
 & \textbf{ResNet-50} & 0.485 & 0.478 & 0.501 & 0.489 \\ \cline{2-6} 
 \multirow{2}{*}{\textbf{PolitiFact}}
 & \textbf{MVAE} & 0.726 & 0.761 & 0.678 & 0.717 \\ 
 & \textbf{att-RNN} & 0.741 & 0.726 & 0.813 & 0.767 \\ 
 & \textbf{SpotFake} & 0.770 & 0.753 & 0.795 & 0.770  \\ 
 & \textbf{EANN} & 0.795 & 0.813 & 0.761 & 0.786  \\ 
 & \textbf{SpotFake+} & 0.856 & 0.878 & 0.846 & 0.862  \\ 
 & \textbf{SAFE} & 0.872 & 0.883 & 0.897 & 0.890  \\ 
 & \textbf{TieFake} & \textbf{0.912} & \textbf{0.931} & \textbf{0.909} & \textbf{0.920}  \\
\hline
\hline
 & \textbf{CNN} & 0.741 & 0.733 & 0.775 & 0.753 \\ 
 & \textbf{GRU} & 0.793 & 0.779 & 0.801 & 0.790 \\ 
 & \textbf{BERT} & 0.836 & 0.872 & 0.829 & 0.850 \\ \cline{2-6} 
 & \textbf{VGG-19} & 0.443 & 0.478 & 0.462 & 0.450 \\ 
 & \textbf{ResNet-50} & 0.454 & 0.469 & 0.458 & 0.463 \\ \cline{2-6} 
 \multirow{2}{*}{\textbf{GossipCop}}
 & \textbf{MVAE} & 0.782 & 0.802 & 0.751 & 0.776 \\ 
 & \textbf{att-RNN} & 0.774 & 0.798 & 0.821 & 0.809 \\ 
 & \textbf{SpotFake} & 0.812 & 0.807 & 0.822 & 0.814  \\ 
 & \textbf{EANN} & 0.833 & 0.842 & 0.835 & 0.838  \\ 
 & \textbf{SpotFake+} & 0.864 & 0.859 & 0.882 & 0.870  \\ 
 & \textbf{SAFE} & 0.831 & 0.843 & 0.894 & 0.868  \\ 
 & \textbf{TieFake} & \textbf{0.892} & \textbf{0.887} & \textbf{0.902} & \textbf{0.894}  \\
\hline
\end{tabular}
}
\end{threeparttable}
}
% \end{adjustbox}
\end{table*}

\subsection{Results}
\label{ssec:results}
Experimental results in Table~\ref{tab::general_performance} further reveal several insightful observations.

(1) In both datasets, multi-domain models work much better than single-domain models. Once again, this validates the benefit of incorporating all kinds of information. For example, additional visual information can be used as complementary information to help detect fake news.

(2) Text-based models work much better than image-based models, demonstrating that textual features could be more helpful than visual ones in detecting fake news. The reason is that it is more challenging to learn the semantic meaning of visual features than textual features. 

(3) SAFE outperforms all baselines on the PolitiFact dataset because SAFE
jointly uses multi-modal (text and visual) and relational information to learn the representation of news, which is more applicable to small sample datasets. In addition, SpotFake and SpotFake+ achieve better results than other baselines on the GossipCop dataset, indicating that the pre-trained BERT and XLNet can obtain better textual information to improve model performance, which is more applicable to big sample datasets.

(4) The proposed \textbf{TieFake} outperforms all the baselines on both large and small sample datasets. In comparison with SAFE on the PolitiFact dataset, our model improves 4.0\% in accuracy, 4.8\% in precision, 1.2\% in recall and 3.0\% in $F_1$. In comparison with SpotFake+ on the GossipCop dataset, our model improves 2.8\% in accuracy, 2.8\% in precision, 2.0\% in recall and 2.4\% in $F_1$. This verifies that the proposed model can jointly capture multi-modal (text and visual) and emotional information, which can better characterize the underlying representation of news content, improving the performance of fake news detection. 

\begin{table}[h]
%\footnotesize
%\scriptsize
    \centering
	\setlength{\belowcaptionskip}{1pt}
	\setlength{\abovecaptionskip}{1pt}
\centering
% \small
\caption{The performance of \textbf{TieFake} variants.}

\label{tab::ablation_performance}
% \begin{adjustbox}{\textwidth}
% \resizebox{\linewidth}{!}
\scalebox{.9}{
\begin{threeparttable}
\setlength{\tabcolsep}{3mm}{
\begin{tabular}{|c|c|c|c|c|c|}
\hline
\textbf{Dataset}&
\textbf{Methods} &
\textbf{Acc} & 
\textbf{Pre} &
\textbf{Rec} & 
\textbf{$\mathbf{F}_1$}\\
\hline
\hline
 & \textbf{TieFake-T} & 0.612 & 0.632 & 0.598 & 0.615 \\ 
 
 & \textbf{TieFake-V} & 0.866 & 0.858 & 0.871 & 0.864 \\ 
 \multirow{1}{*}{\textbf{PolitiFact}}
 & \textbf{TieFake-E} & 0.857 & 0.851 & 0.878 & 0.864
 \\ 
 & \textbf{TieFake-S} & 0.904 & 0.925 & 0.902 & 0.914 \\
 & \textbf{TieFake} & \textbf{0.912} & \textbf{0.931} & \textbf{0.909} & \textbf{0.920}  \\
\hline
\hline
 & \textbf{TieFake-T} & 0.593 & 0.588 & 0.623 & 0.605 \\ 
 
 & \textbf{TieFake-V} & 0.862 & 0.872 & 0.846 & 0.859 \\ 
 \multirow{1}{*}{\textbf{GossipCop}}
 & \textbf{TieFake-E} & 0.854 & 0.869 & 0.851 & 0.860 \\
 & \textbf{TieFake-S} & 0.886 & 0.891 & 0.895 & 0.892 \\
 & \textbf{TieFake} & \textbf{0.892} & \textbf{0.887} & \textbf{0.902} & \textbf{0.894}  \\
\hline
\end{tabular}
}
\end{threeparttable}
}
% \end{adjustbox}
\end{table}
\subsection{Ablation study}
\label{ssec:ablation}
The proposed \textbf{TieFake} contains several components, thus we additionally compare the variants of \textbf{TieFake} to show the impact of each component.  

\begin{itemize}
    \item \textbf{TieFake-T}: A variant of TieFake with the textual information is removed;
    
    \item \textbf{TieFake-V}: A variant of TieFake with the visual information is removed;
    
    \item \textbf{TieFake-E}: A variant of TieFake with the author potential emotion is removed.

    \item \textbf{TieFake-S}: A variant of TieFake with the similarity between the title and text is removed.
    
\end{itemize}
Results in Table ~\ref{tab::ablation_performance} indicate that (1) integrating news' textual information, visual information, and author's potential emotion improves the performance; (2) the proposed method \textbf{TieFake} outperforms \textbf{TieFake-E}, which shows the effectiveness of introducing the potential emotion to our model; (3) the proposed method \textbf{TieFake} outperforms \textbf{TieFake-S}, which proves that mining the similarity between the title and the text is effective for detecting fake news; (4) textual information is much more critical than visual information and potential author emotion.

\section{Conclusion}
\label{sec:conclusion}

In this paper, we propose a simple but effective framework named \textbf{TieFake} to detect fake news, which utilizes both textual and visual features of news content and investigates subjective emotional features of authors. To our knowledge, it is the first model to exploit author emotion in the multi-modal fake news detection task. Primarily, we propose a novel attention mechanism to learn the similarity between the title and text. Experimental results verify the effectiveness of our model. The proposed method can be extended in the future to consider more complex information, e.g., network and video information. Additionally, there is still room for improvement on more complex fusion techniques to understand how different modalities play a role in fake news detection.

\section{Acknowledgment}

This work was supported by the National Natural Science Foundation of China (No. 62276053) and the Sichuan Science and Technology Program (No. 22ZDYF3621).

%\newpage
\bibliographystyle{IEEEtran}
\bibliography{spkformer}

\end{document}